%% file: ieee_its_journal.tex
\documentclass[journal]{IEEEtran}

\usepackage{soul}
\usepackage[dvipsnames]{xcolor}
\usepackage{amsmath}
\usepackage[pdftex]{graphicx}
\usepackage{hyperref}
\usepackage{subcaption}
\DeclareMathOperator*{\argmin}{arg\,min}
\usepackage{multirow}

\begin{document}
\bstctlcite{IEEEexample:BSTcontrol}

\title{Scene-Graph Augmented Data-Driven Risk Assessment of Autonomous Vehicle Decisions}

\author{
    \IEEEauthorblockN{Shih-Yuan Yu, Arnav V. Malawade, Deepan Muthirayan, Pramod~P.~Khargonekar, 
    Mohammad A. Al Faruque}\\
    \IEEEauthorblockA{Department of Electrical Engineering \& Computer Science, University of California, Irvine CA, USA.}
    \IEEEauthorblockA{\\\{shihyuay,malawada,dmuthira,pramod.khargonekar,alfaruqu\}@uci.edu }
}

\maketitle

\input{sections/0_abstract}


\input{sections/1_introduction}
\input{sections/2_related_works}
\input{sections/3_approach}
\input{sections/4_experiments}
\input{sections/5_conclusion}
\input{sections/6_acknowledgement}

\ifCLASSOPTIONcaptionsoff
  \newpage
\fi

\bibliographystyle{IEEEtran}
\bibliography{IEEEabrv, references}

\input{sections/7_IEEE_author_bios}




\end{document}

%% file: sections/0_abstract.tex
\begin{abstract}
Despite impressive advancements in Autonomous Driving Systems (ADS), navigation in complex road conditions remains a challenging problem. 
There is considerable evidence that evaluating the subjective risk level of various decisions can improve ADS' safety in both normal and complex driving scenarios. 
However, existing deep learning-based methods often fail to model the relationships between traffic participants and can suffer when faced with complex real-world scenarios.
Besides, these methods lack \textit{transferability} and \textit{explainability}.
To address these limitations, we propose a novel data-driven approach that uses \textit{scene-graphs} as intermediate representations.
Our approach includes a Multi-Relation Graph Convolution Network, a Long-Short Term Memory Network, and attention layers for modeling the subjective risk of driving maneuvers.
To train our model, we formulate this task as a supervised scene classification problem. 
We consider a typical use case to demonstrate our model's capabilities: lane changes.
We show that our approach achieves a higher classification accuracy than the state-of-the-art approach on both large (96.4\% vs. 91.2\%) and small (91.8\% vs. 71.2\%) synthesized datasets, also illustrating that our approach can learn effectively even from smaller datasets. 
We also show that our model trained on a synthesized dataset achieves an average accuracy of 87.8\% when tested on a real-world dataset compared to the 70.3\% accuracy achieved by the state-of-the-art model trained on the same synthesized dataset, showing that our approach can more effectively transfer knowledge.
Finally, we demonstrate that the use of spatial and temporal attention layers improves our model's performance by 2.7\% and 0.7\% respectively and increases its explainability.
\end{abstract}

%% file: sections/1_introduction.tex
\section{Introduction}
\label{sec:introduction}
\begin{figure*}[!ht]
    \centering
    \includegraphics[width=1.\linewidth]{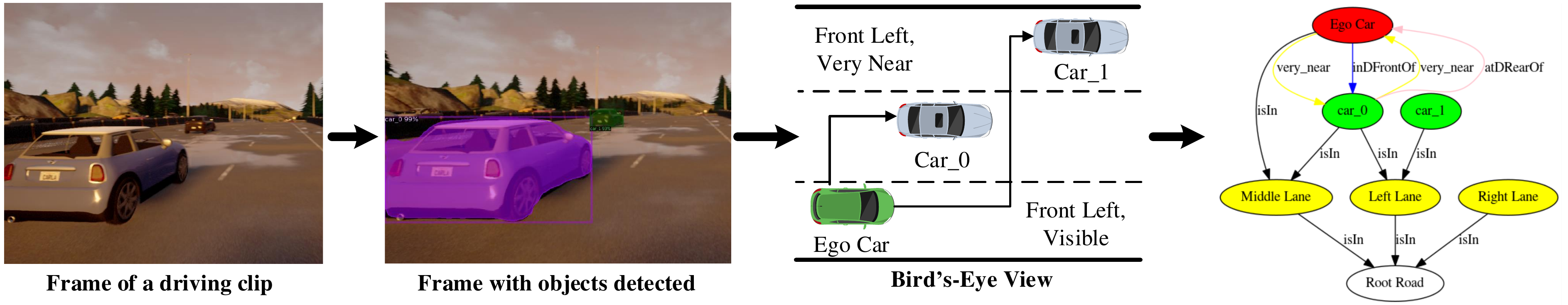}
    \caption{An illustration of \textit{scene-graph} extraction using the \textit{Real Image Pipeline}. In this process, the first step is to detect a list of objects on each frame of a clip. 
    Then, we project each frame to its bird's-eye view to better approximate the spatial relations between objects.
    Finally, we construct a \textit{scene-graph} using the list of detected objects and their attributes.}
    \label{fig:network_archi}
\end{figure*}
Autonomous Driving Systems (ADSs) have advanced significantly in recent years. 
However, navigation is still challenging in complex urban environments since the scenarios are highly variable and complex~\cite{montgomery2018america, yurtsever2019survey, paden2016survey}.
The continued reports of autonomous vehicle crash only highlight these challenges~\cite{lavrinc2014bad, davies2016google, mcfarland2016, tblee2019}.
A risk-based approach for autonomous driving has the potential to address this challenge and better assure driving safety.
Within this context, the effectiveness of understanding the driving scenes and quantifying the risk of driving decisions becomes particularly crucial for ADSs.

In prior research,  the problem of risk assessment for autonomous driving has been tackled by modeling either the \textit{objective risk} or the \textit{subjective risk}~\cite{grayson2003risk, fuller2005towards, bao2019personalized}.
The \textit{objective risk} is defined as the objective probability of an accident occurring and is usually determined by statistical analysis~\cite{grayson2003risk}. 
Some works have focused on minimizing the \textit{objective risk} by modeling the trajectories of vehicles~\cite{6025208, 8569759} to guarantee safe driving.

\textit{Subjective risk} refers to the driver's own perceived risk and is an output of the driver's cognitive process~\cite{fuller2005towards, bao2019personalized}.
Studies suggest that modeling the \textit{subjective risk} can be a better option as drivers typically avoid taking a risky driving action based on their subjective perception~\cite{bao2019personalized}. 
For this reason, our objective in this work is to build a model for subjective risk assessment.

Several papers have leveraged state-of-the-art deep learning architectures for modeling subjective risk~\cite{yurtsever2019survey, yurtsever2019risky}.
Such methods typically apply Convolutional Neural Networks (CNNs) and Long-Short Term Memory Networks (LSTMs) and have been proven to be effective at capturing features essential for modeling subjective risk in both spatial and temporal domains~\cite{yurtsever2019risky}.
However, it is unclear whether these methods can capture critical higher-level information, such as the relationships between the traffic participants in a given scene.
Failure in capturing these relationships can result in poor ADS performance in complex scenarios.

Also, training these networks requires large datasets covering a wide range of ``corner cases'' (especially risky driving scenarios), which are expensive and time-consuming to generate~\cite{dosovitskiy2017carla}. 
Many researchers resort to using synthesized datasets containing many examples of these corner cases to address this issue. 
However, for these to be valuable, a model must be able to transfer the knowledge gained from simulated training data to real-world situations.
A standard method for measuring a model's ability to generalize is \textit{transferability}, where a model's accuracy on a dataset different from the training dataset is evaluated. 
If a model can transfer the knowledge gained from a simulated training set to a real-world testing set effectively, it is likely that it will perform better in unseen real-world scenarios.

Even if these existing methods can transfer knowledge well, the predictions of such methods lack \textit{explainability}, which is crucial for establishing trust between ADSs and human drivers~\cite{bao2019personalized, bahdanau2014neural, adadi2018peeking}. 
\textit{Explainability} refers to the ability of a model to effectively communicate the factors that influenced its decision-making process for a given input, particularly those that might lead the model to make incorrect decisions~\cite{adadi2018peeking, knyazev2019understanding}. 
If a model can give attention to the aspects or entities in a traffic scene that make the scenario risky or non-risky, it can improve its decision and its decisions become more explainable~\cite{vaswani2017attention}.

Overall, designing a risk assessment system for ADSs using data-driven approaches presents the following challenges:
\begin{itemize}
    \item Designing a reliable method that can handle a wide range of complex and unpredictable traffic scenarios. 
    \item Building a model that is transferable from the simulation setting to the real-world setting because the real-world datasets for supervised training are limited.
    \item Building a model that can provide explainable decisions.
\end{itemize}

In this paper, we present our work on the modeling of \textit{subjective risk} for a specific driving maneuver: lane change.
This task by itself is crucial, given that 7.62\% of all traffic accidents between light vehicles can be attributed to improper execution of lane change~\cite{najm2007pre}. 
We propose a \textit{scene-graph} augmented data-driven approach for assessing the subjective risk of driving maneuvers, where the \textit{scene-graphs} serve as intermediate representations (IR) as shown in Figure~\ref{fig:network_archi}. 
The key advantage is that using \textit{scene-graphs} as IRs allows us to model the relationships between the participants in a traffic scene, thus potentially improving the model's understanding of a scene. 

Our architecture consists of three major components: (i) a pipeline to convert the images of a driving clip to a sequence of \textit{scene-graphs}, (ii) a Multi-Relational Graph Convolution Network (MR-GCN) to convert each of the \textit{scene-graphs} to an embedding (a vectorized representation), and (iii) an LSTM for temporally modeling the sequence of embeddings of the respective \textit{scene-graphs}.
Our model also contains multiple attention layers: (i) a node attention layer before the embedding of a \textit{scene-graph} is computed, and (ii) an attention layer on top of the LSTM, both of which can further improve its performance and the explainability.
For training the model, we propose formulating the problem of subjective risk assessment as a supervised \textit{scene-graph} sequence classification problem. 

Our key contributions are as follows:
\begin{itemize}
    \item We present a novel \textit{scene-graph} augmented data-driven approach for assessing the risk of driving actions in autonomous vehicles. 
    \item We demonstrate that our approach outperforms existing methods at risk assessment across a wide range of scenarios using lane change as a use case.
    \item We demonstrate that the use of multi-level attention in our proposed approach provides better explainability.
    \item We demonstrate that our \textit{scene-graph} based approach can better transfer knowledge gained from simulated environments to real-world risk assessment tasks.
\end{itemize}

The rest of the paper is structured as follows:
In Section~\ref{sec:related_work}, we discuss related works. In Section~\ref{sec:approach} we introduce our \textit{scene-graph} augmented approach. 
In Section~\ref{sec:experimental_results} we discuss our experimental results. 
Finally, in Section~\ref{sec:conclusion} we present our conclusions.

%% file: sections/2_related_works.tex
\section{Related Work}
\label{sec:related_work}
\subsection{Design Philosophies in ADSs}
Two broad approaches for designing ADSs are (i)  modular design , (ii)  end-to-end design~\cite{yurtsever2019survey}. 
Most modular approaches comprise a pipeline of separate components from the sensory inputs to the actuator outputs, while end-to-end approaches generate output directly from their sensory inputs~\cite{pomerleau1989alvinn, bojarski2016end}.
One advantage of a modular design approach is the division of a task into an easier-to-solve set of sub-tasks that have been addressed in other fields such as robotics~\cite{laumond1998robot}, computer vision~\cite{jain1995machine} and vehicle dynamics~\cite{rajamani2011vehicle}. 
Therefore, prior knowledge from these fields can be leveraged when designing the components corresponding to the sub-tasks. 
However, one disadvantage of such an approach is the complexity of the whole pipeline~\cite{yurtsever2019survey}. 
End-to-end approaches can achieve good performance with a smaller network size because they perform feature extraction from sensor inputs implicitly through the network's hidden layers~\cite{bojarski2016end}.
However, the authors in~\cite{chen2015deepdriving} point out that the needed level of supervision is too weak for the end-to-end model to learn critical controlling information (e.g., from image to steering angle), so it can fail to handle complicated driving maneuvers.

A third approach was first proposed by DeepDriving~\cite{chen2015deepdriving}, called the \textit{direct perception} approach. 
In their approach, a set of \textit{affordance indicators}, such as the distance to the lane markings and the distance to cars in the current and adjacent lanes, are extracted from an image and serves as an IR for generating the final control output.
They prove that the use of this IR is effective for simple driving tasks such as lane following and for generalizing to real-world environments.
Authors in~\cite{bansal2018chauffeurnet} use a collection of filtered images, each representing a piece of distinct information, as the IR.
They state that the IR used in their approach allows the training to be conducted on real or simulated data, facilitating testing and validation in simulations before testing on a real car.
Moreover, they show that it is easier to synthesize perturbations to the driving trajectory at the mid-level representations than at the level of raw sensors, enabling them to produce non-expert behaviors such as off-road driving and collisions.
The authors in~\cite{yurtsever2019risky} use Mask-RCNN~\cite{he2017mask} to color the vehicles in each input image, producing a form of IR. 
In contrast to the works mentioned above, our approach uses a \textit{scene-graph} IR that encodes the spatial and semantic relations between all the traffic participants in a frame.

\subsection{Graph-Based Driving Scene Understanding}

Several papers have applied graph-based formulations for driving scene understanding~\cite{mylavarapu2020towards, li2019learning, mylavarapu2020understanding}. 
In~\cite{li2019learning}, the authors propose a 3D-aware egocentric Spatio-temporal interaction framework that uses both an \textit{Ego-Thing} graph and an \textit{Ego-Stuff} graph, which encode how the ego vehicle interacts with both moving and stationary objects in a scene, respectively.
In~\cite{mylavarapu2020towards, mylavarapu2020understanding}, the authors propose a pipeline using a multi-relational graph convolutional network (MR-GCN) for classifying the driving behaviors of traffic participants.
The MR-GCN is constructed by combining spatial and temporal information, including relational information between moving objects and landmark objects.
Our work is primarily inspired by~\cite{mylavarapu2020towards, mylavarapu2020understanding} but differs in the application and network architecture.
These papers focus on predicting each object's behavior in the \textit{scene-graph} while our work focuses on assessing the subjective risk of the entire scenario.
Consequently, we propose a network architecture that implements more components such as node-attention, graph pooling layers, and readout operations.

\subsection{Risk Assessment}
\label{subsec:risk}
Several works have studied subjective risk assessment for autonomous driving systems~\cite{fuller2005towards, bao2019personalized, yurtsever2019risky, yurtsever2018integrating}. 
In~\cite{yurtsever2018integrating}, Hidden Markov Models (HMMs) and Language Models are used to detect unsafe lane change events.
The approach taken in~\cite{yurtsever2019risky} is the most related to our work as it infers the risk-level of overall driving scenes with a deep Spatio-temporal neural network architecture.
By using Mask-RCNN~\cite{he2017mask} to generate an IR for each image, their approach achieves a 3\% performance gain in risk assessment. 
They show that the architecture with Semantic Mask Transfer (SMT) + CNN + LSTM can perform 25\% better than the architecture with Feature Transfer (FT) + Frame-by-Frame (FbF), indicating that capturing the spatial and temporal features just from a monocular camera can be useful in modeling subjective risk.
However, this approach only considers the spatial features (the latent vector output of the CNN layers) of a frame instead of the relations between all the traffic participants. 
Our work uses \textit{scene-graphs} as IRs to capture the high-level relationships between all the traffic participants of a scene.

%% file: sections/3_approach.tex
\section{Scene-Graph Augmented Approach for Risk Assessment}
\label{sec:approach}
\begin{figure*}[!ht]
    \includegraphics[width=1.0\linewidth]{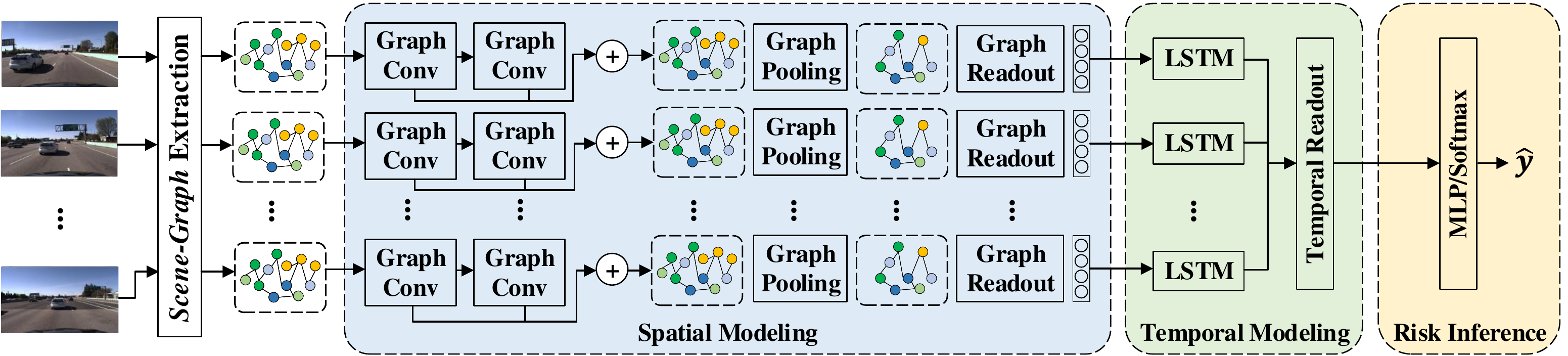}
    \caption{An illustration of our model's architecture. 
    It first converts each image $I_t \in \mathbf{I}$ to a \textit{scene-graph} $G_t$ via the \textit{Scene-Graph Extraction Pipeline}. Next, it converts each $G_t$ to its corresponding \textit{scene-graph} embedding $\mathbf{h}_{G_t}$ with layers in Spatial Modeling. Then, it temporally models these \textit{scene-graph} embeddings to acquire the spatio-temporal representation $\mathbf{Z}$ for a \textit{scene-graph} sequence. Finally, the risk inference $\hat{y}$ of a clip is calculated from $\mathbf{Z}$ using an MLP with a \textit{Softmax} activation function.}
    \label{fig:archi}
\end{figure*}

In this section, we discuss our proposed approach for \textit{scene-graph} augmented risk assessment.
In our work, we make the same assumption used in~\cite{yurtsever2019risky} that the set of driving sequences can be partitioned into two jointly exhaustive and mutually exclusive subsets: risky and safe. 
We denote the sequence of images of length $T$ by $\mathbf{I} = \{I_1, I_{2}, I_{3}, ..., I_T\}$.
We assume the existence of a spatio-temporal function $f$ that outputs whether a sequence of driving actions $x$ is safe or risky via a risk label $y$, as given in Equation~\ref{formular1-v2}.
\begin{equation}
\label{formular1-v2}
    y = f(\mathbf{I}) = f(\{I_1, I_2, I_3, ..., I_{T-1}, I_T\}),
\end{equation} 
where 
\begin{equation}
    y=\left \{
    \begin{array}{ll}
         (1,0), &  \text{if the driving sequence is safe}\\
         (0,1), &  \text{if the driving sequence is risky}.
    \end{array}
    \right.
\end{equation}

In this section, we propose a suitable model for approximating the function $f$.
In the model we propose, the first step is the extraction of the \textit{scene-graph} $G_t$ from each image $I_t$ of the video clip $\mathbf{I}$.
This is achieved by a series of processes that we collectively call the \textit{Scene-Graph Extraction Pipeline}, described in Section~\ref{subsec:extraction}.
In the second step, these \textit{scene-graphs} are passed through graph convolution layers and an attention-based graph pooling layer. 
The graph-level embeddings of each \textit{scene-graph}, $\mathbf{h}_{G_t}$, are then calculated using a graph readout operation.
Next, these \textit{scene-graph} embeddings are passed sequentially to LSTM cells to acquire the \textit{spatio-temporal} representation, denoted as $\mathbf{Z}$, of each \textit{scene-graph} sequence.
Lastly, we use a Multi-Layer Perceptron (MLP) layer with a \textit{Softmax} activation function to acquire the final inference, denoted as $\hat{y}$, of the risk for each driving sequence $\mathbf{I}$.
We describe more details regarding each of these components of our model in Section~\ref{subsec:model}.

\subsection{Scene-Graph Extraction Pipeline}
\label{subsec:extraction}
Several approaches have been proposed for extracting scene-graphs from images by detecting the objects in a scene and then identifying their visual relationships~\cite{xu2017scene, yang2018graph}. 
However, these works have focused on single general images instead of a sequence of images as it arises in autonomous driving, where higher accuracy is demanded. 
Thus, we adopted a partially rule-based process to extract objects and their attributes from images called the \textit{Real Image Pipeline}.
Besides, to evaluate how our approach performs with \textit{scene-graphs} containing ground truth information, we use the \textit{Carla Ground Truth (GT) Pipeline} as a surrogate for the ideal situation where the attributes for each object can be correctly extracted.
After the objects in a scene and their attributes have been extracted, the \textit{scene-graphs} are constructed as described in \ref{subsec:constructgraph}.
We discuss these two approaches in detail below.

\subsubsection{Real Image Pipeline}
\label{subsec:real}
In this pipeline, object attributes and bounding boxes are extracted directly from images using state-of-the-art image processing techniques.
As Figure~\ref{fig:network_archi} shows, we first convert each image $I_t$ into a collection of objects $O_t$ using Detectron2, a state-of-the-art object detection model based on Faster RCNN~\cite{wu2019detectron2, ren2015faster}.
Next, we use OpenCV's perspective transformation library to generate a top-down perspective of the image, commonly known as a "birds-eye view" projection~\cite{opencv_library}. 
This projection allows us to approximate each object's location relative to the road markings and the ego vehicle.
Next, for each detected object in $O_t$, we use its estimated location and class type (cars, motorcycles, pedestrians, lanes, etc.) to compute the attributes required in building the \textit{scene-graph}.

\subsubsection{Carla Ground Truth Pipeline} 
\label{subsec:carla}
Object detection and location estimation with solely monocular camera can be unstable because of factors such as weather and camera position~\cite{caesar2020nuscenes}, which can impact the correctness of our image-based \textit{scene-graph} construction pipeline and thus our approach's performance. 
To evaluate our methodology under the assumption that object attributes can be extracted without error, we build our \textit{scene-graphs} using the ground-truth location and class information for each vehicle in the \textit{Carla GT Pipeline}. 
We extract this information directly from Carla simulator~\cite{dosovitskiy2017carla} without using any image processing steps.

\subsubsection{Graph Construction} 
\label{subsec:constructgraph}
After collecting the list of objects in each image and their attributes, we begin constructing the corresponding \textit{scene-graphs} as follows.
For each image $I_t$, we denote the corresponding \textit{scene-graph} by $G_t = \{O_t, A_t\}$ and model it as a directed multi-graph where multiple types of edges connect nodes.
The nodes of a \textit{scene-graph}, denoted as $O_t$, represent the objects in a scene such as lanes, roads, traffic signs, vehicles, pedestrians, etc. 
The edges of a \textit{scene-graph} are represented by the corresponding adjacency matrix $A_t$, where each value in $A_t$ represents the type of the edges. 
The edges between two nodes represent the different kinds of relations between them (e.g., near, Front\_Left, isIn, etc.).

In assessing the risk of driving behaviors, traffic participants' relations that we consider to be useful are the distance relations and the directional relations.
The assumption made here is that the local proximity and positional information of one object will influence the other's motion only if they are within a certain distance.
Therefore, in this work, we extract only the location information for each object and adopt a simple rule to determine the relations between the objects using their attributes (e.g., relative location to the ego car), as shown in Figure~\ref{fig:network_archi}. 
For distance relations, we assume two objects are related by one of the relations $r\in$ \{\textit{Near\_Collision} (4 ft.), \textit{Super\_Near} (7 ft.), \textit{Very\_Near} (10 ft.), \textit{Near} (16 ft.), \textit{Visible} (25 ft.)\} if the objects are physically separated by a distance that is within that relation's threshold.
In the case of the directional relations, we assume two objects are related by the relation $r \in$ \{\textit{Front\_Left}, \textit{Left\_Front}, \textit{Left\_Rear}, \textit{Rear\_Left}, \textit{Rear\_Right}, \textit{Right\_Rear}, \textit{Right\_Front}, \textit{Front\_Right}\} based on their relative positions if they are within the \textit{Near} threshold distance from one another.

In addition to directional and distance relations, we also implement the \textit{isIn} relation that connects vehicles with their respective lanes.
For the \textit{Carla GT Pipeline}, we extract the ground-truth lane assignments for each vehicle from the simulator directly.
For the \textit{Real Image Pipeline}, we use each vehicle's horizontal displacement relative to the ego vehicle to assign vehicles to either the \textit{Left Lane}, \textit{Middle Lane}, or \textit{Right Lane} based on a known lane width. 
Our abstraction only includes these three-lane areas and, as such, we map vehicles in all left lanes to the same \textit{Left Lane} node and all vehicles in right lanes to the \textit{Right Lane} node. 
If a vehicle overlaps two lanes (i.e., during a lane change), we assign it an \textit{isIn} relation to both lanes.
Figure~\ref{fig:network_archi} illustrates an example of resultant \textit{scene-graph}.

\subsection{Scene-graph Sequence Based Risk Assessment Model}
\label{subsec:model}

The model we propose consists of three major components: spatial model, temporal model, and risk inference.
The spatial model outputs the embedding $h_{G_t}$ for each scene-graph $G_t$.
The temporal model processes the sequence of \textit{scene-graph} embeddings $\mathbf{h}_I = \{h_{G_1}, h_{G_2}, ..., h_{G_T}\}$ and produces the spatio-temporal embedding $\mathbf{Z}$. 
The risk inference component outputs each driving clip's final risk assessment, denoted as $\hat{Y}$, by processing the Spatio-temporal embedding $\mathbf{Z}$.
The overall network architecture is shown in Figure~\ref{fig:archi}. We discuss each of these components in detail below.

\subsubsection{Spatial Modeling}
The spatial model we propose uses MR-GCN layers to compute the embedding for a \textit{scene-graph}. 
The use of MR-GCN allows us to capture multiple types of relations on each \textit{scene-graph} $G_t = \{O_t, A_t\}$.
In the \textit{Message Propagation} phase, a collection of node embeddings and their adjacency information serve as the inputs to the MR-GCN layer. 
Specifically, the $l$-th MR-GCN layer updates the node embedding, denoted as $\mathbf{h}^{(l)}_v$, for each node $v$ as follows:
\begin{equation}
    \mathbf{h}^{(l)}_v = \mathbf{\Phi}_{\textrm{0}} \cdot
        \mathbf{h}^{(l-1)}_v + \sum_{r \in \mathbf{A_{t}}} \sum_{u \in \mathbf{N}_r(v)}\frac{1}{|\mathbf{N}_r(v)|} \mathbf{\Phi }_r \cdot \mathbf{h}^{(l-1)}_u,
\end{equation}
where $N_r(v)$ denotes the set of neighbor indices of node $v$ with the relation $r\in A_{t}$. 
$\mathbf{\Phi }_r$ is a trainable relation-specific transformation for relation $r$ in MR-GCN layer. 
Since the information in $(l-1)$-th layer can directly influence the representation of the node at $l$-th layer, MR-GCN uses another trainable transformation $\mathbf{\Phi }_0$ to account for the self-connection of each node using a special relation~\cite{schlichtkrull2018modeling}. 
Here, we initialize each node embedding $\mathbf{h}^{(0)}_{v}$, $\forall v \in O_t$, by directly converting the node's type information to its corresponding one-hot vector.

Typically, the node embedding becomes more refined and global as the number of graph convolutional layers, $L$, increases. 
However, the authors in~\cite{xu2018powerful} also suggest that the features generated in earlier iterations might generalize the learning better. 
Therefore, we consider the node embeddings generated from all the MR-GCN layers. 
To be more specific, we calculate the embedding of node $v$ at the final layer, denoted as $\mathbf{H}^{L}_v$, by concatenating the features generated from all the MR-GCN layers, as follows,
\begin{equation}
\mathbf{H}^{L}_v = \textbf{CONCAT}(\{\mathbf{h}^{(l)}_v\}|l=0, 1, ...,L).
\end{equation}
We denote the collection of node embeddings of \textit{scene-graph} $G_t$ after passing through $L$ layers of MR-GCN as $\mathbf{X}^{prop}_{t}$ ($L$ can be 1, 2 or 3).

The node embedding $\mathbf{X}^{prop}_{t}$ is further processed with an attention-based graph pooling layer.
As stated in~\cite{knyazev2019understanding}, such an attention-based pooling layer can improve the explainability of predictions and is typically considered as a part of a unified computational block of a graph neural network (GNN) pipeline.
In this layer, nodes are pooled according to the scores predicted from either a trainable simple linear projection~\cite{gao2019graph} or a separate trainable GNN layer~\cite{lee2019self}.
We denote the graph pooling layer that uses the \textbf{SCORE} function in~\cite{gao2019graph} as \textit{TopkPool} and the one that uses the \textbf{SCORE} function in~\cite{lee2019self} as \textit{SAGPool}.
The calculation of the overall process is presented as follows: 
\begin{equation}
\mathbf{\alpha} = \textbf{SCORE}(\mathbf{X}^{prop}_{t}, \mathbf{A_{t}}),
\end{equation}
\begin{equation}
\label{eq:pool}
\mathbf{P} = \mathrm{top}_k(\mathbf{\alpha}), 
\end{equation}
where $\mathbf{\alpha}$ stands for the coefficients predicted by the graph pooling layer for nodes in $G_t$ and $\mathbf{P}$ represents the indices of the pooled nodes which are selected from the top $k$ of the nodes ranked according to $\alpha$. The number $k$ of the nodes to be pooled is calculated by a pre-defined pooling ratio, $pr$, and using $k = pr \times |O_t|$, where we consider only a constant fraction $pr$ of the embeddings of the nodes of a scene-graph to be relevant (i.e., 0.25, 0.5, 0.75). 
We denote the node embeddings and edge adjacency information after pooling by $\mathbf{X}^{pool}_{t}$ and $\mathbf{A^{pool}_{t}}$ and are calculated as follows:
\begin{equation}
\mathbf{X}^{pool}_{t}= (\mathbf{X}^{prop}_{t} \odot\mathrm{tanh}(\mathbf{\alpha}))_{\mathbf{P}}, 
\end{equation}
\begin{equation}
\mathbf{A^{pool}_{t}} = \mathbf{A^{prop}_{t}}_{(\mathbf{P},\mathbf{P})}. \\
\end{equation}
where $\odot$ represents an element-wise multiplication, $()_{\mathbf{P}}$ refers to the operation that extracts a subset of nodes based on $P$ and  $()_{(\mathbf{P},\mathbf{P})}$ refers to the formation of the adjacency matrix between the nodes in this subset.

Finally, our model aggregates the node embeddings of the graph pooling layer, $\mathbf{X}^{pool}_{t}$, using a graph \textbf{READOUT} operation, to produce the final graph-level embedding $\mathbf{h}_{G_t}$ for each \textit{scene-graph} $G_t$ as given by
\begin{equation}
    \mathbf{h}_{G_t} = \textbf{READOUT}(\mathbf{X}^{pool
    }_{t}),
\end{equation}
where the \textbf{READOUT} operation can be either summation, averaging, or selecting the maximum of each feature dimension, over all the node embeddings, known as \textit{sum-pooling}, \textit{mean-pooling}, or \textit{max-pooling}, respectively.
The process until this point is repeated across all images in $\mathbf{I}$ to produce the sequence of embedding, $\mathbf{h}_I$.

\subsubsection{Temporal Modeling}
\label{subsec:temporal_modeling}
The temporal model we propose uses an LSTM for converting the sequence of scene-graph embeddings $\mathbf{h}_I$ to the combined spatio-temporal embedding $\mathbf{Z}$. 
For each timestamp $t$, the LSTM updates the hidden state $p_t$ and cell state $c_t$ as follows,
\begin{equation}
    p_t, c_t = \mathbf{LSTM}(\mathbf{h}_{G_t}, c_{t-1}),
\end{equation}
where $\mathbf{h}_{G_t}$ is the final \textit{scene-graph} embedding from timestamp $t$.
After the LSTM processes all the scene-graph embeddings, a temporal readout operation is applied to the resultant output sequence to compute the final Spatio-temporal embedding $Z$ given by
\begin{equation}
    \mathbf{Z} = \textbf{TEMPORAL\_READOUT}(p_1, p_2, ..., p_T)
\end{equation}
where the $\textbf{TEMPORAL\_READOUT}$ operation could be extracting only the last hidden state $p_T$ (LSTM-last), or be a temporal attention layer (LSTM-attn).

In~\cite{bahdanau2014neural}, adding an attention layer $b$ to the encoder-decoder based LSTM architecture is shown to achieve better performance in Neural Machine Translation (NMT) tasks. For the same reason, we include \textit{LSTM-attn} in our architecture.
\textit{LSTM-attn} calculates a context vector $q$ using the hidden state sequence $\{p_1, p_2, ..., p_T\}$ returned from the LSTM encoder layer as given by 
\begin{equation}
\label{eq:temporalattn}
q = \sum^{T}_{t=1} \beta_t p_t
\end{equation}
where the probability $\beta_t$ reflects the importance of $p_t$ in generating $q$.
The probability $\beta_t$ is computed by a {\it Softmax} output of an energy function vector $e$, whose component $e_t$ is the energy corresponding to $p_t$. Thus, the probability $\beta_t$ is formally given by 
\begin{equation}
    \beta_t = \frac{\text{exp}(e_t)}{\sum_{k=1}^T \text{exp}(e_k)},
\end{equation}
where the energy $e_t$ associated with $p_t$ is given by $e_t = b(s_0, p_t)$. 
The temporal attention layer $b$ scores the importance of the hidden state $p_t$ to the final output, which in our case is the risk assessment.
The variable $s_0$ in the temporal attention layer $b$ is computed from the last hidden representation $p_T$. The final Spatio-temporal embedding for a video clip, $Z$, is computed by feeding the context vector $q$ to another LSTM decoder layer.

\subsubsection{Risk Inference} 
The last piece of our model is the risk inference component that computes the risk assessment prediction $\hat{Y}$ using the spatio-temporal embedding $\mathbf{Z}$. This component is composed of a MLP layer followed by a \textit{Softmax} activation function. Thus, the prediction $\hat{Y}$ is given by
\begin{equation}
    \hat{Y} = Softmax(\mathbf{MLP}(Z))
\end{equation}
The loss for the prediction is calculated as follows,
\begin{equation}
    \argmin \mathbf{CrossEntropyLoss}(Y, \hat{Y})
\end{equation}
For training our model, we use a mini-batch gradient descent algorithm that updates its parameters by training on a batch of \textit{scene-graph} sequences. 
To account for label imbalance, we apply class weighting when calculating loss.
Besides, several dropout layers are inserted into the network to reduce overfitting.

%% file: sections/4_experiments.tex
\section{Experimental Results}
\label{sec:experimental_results}
In this section, we provide extensive experimental results to illustrate the accuracy of our model and the ability of our model to transfer knowledge (transferability). 
We do this by providing comparisons between our model and a state-of-the-art SMT+CNN+LSTM based risk assessment model~\cite{yurtsever2019risky}; we refer to this model as the \textit{Baseline}.
Besides, we provide results for our model's best hyper-parameter setting and perform an ablation study to evaluate the contribution of each major component in our model. 

\subsection{Dataset Preparation}
\label{subsec:dataset}
We prepared two types of datasets for our experiments (i) real-world driving datasets and (ii) synthesized datasets. We generated the real-world dataset by extracting lane change clips from the Honda Driving Dataset (HDD)~\cite{Ramanishka_behavior_CVPR_2018}.
To create the synthesized datasets, we developed a tool to generate lane changing clips using the Carla\footnote{\url{https://github.com/carla-simulator/carla}} and Carla Scenario Runner\footnote{\url{https://github.com/carla-simulator/scenario_runner}}.

Carla is an open-source driving simulator~\cite{dosovitskiy2017carla} that allows users to control a vehicle in either manual mode or autopilot mode.
The Carla Scenario Runner contains a set of atomic controllers that enable users to control a car in a driving scene and perform complex driving maneuvers.
We modified the user script in Carla so that it can (i) select one autonomous car randomly and switch its mode to manual mode, and then (ii) utilize Scenario Runner's function to force the vehicle to change lanes.

The data generating tool allows us to fabricate lane changing clips directly instead of extracting them from long driving clips. 
We generated a wide range of simulated lane changes using the various presets in Carla that allowed us to specify the number of cars, pedestrians, weather and lighting conditions, etc.
Also, through the APIs provided by the Traffic Manager (TM) of the Carla simulator, we were able to customize the driving characteristics of every autonomous vehicle, such as the intended speed considering the current speed limit, the chance of ignoring the traffic lights, or the chance of neglecting collisions with other vehicles.
Overall, this allowed us to simulate a wide range of very realistic urban driving environments and generate synthesized datasets suitable for training and testing a model.

To label the lane change clips in both the real-world and synthesized datasets, we performed an annotation process similar to the one used in~\cite{yurtsever2019risky}.
First, in this process, human annotators were asked to assign a risk score to each clip that ranges from -2 and 2, where 2 implies a highly risky lance change and -2 means the safest lane change.
Then, for each lane change clip, all the risk labels of annotators were averaged and converted to the binary label $y$ as follows: if the average is $\leq$ 0, then assign $y=0$ (safe); else assign $y=1$ (risky).

We generated two synthesized datasets: a \textit{271-syn} dataset and a \textit{1043-syn} dataset, containing 271 and 1,043 lane changing clips, respectively.
In addition, we sub-sampled the \textit{271-syn} and \textit{1043-syn} datasets further to create two balanced datasets that have a 1:1 distribution of risky to safe lane changes: \textit{96-syn} and \textit{306-syn}.
We call the real-driving dataset as \textit{571-honda} as it contains 571 lane changing video clips.

We randomly split each dataset into a training set and a testing set by the ratio 7:3 such that the split is stratified, i.e., the proportion of risky to safe lane change clips in each of the splits is the same. 
The models are evaluated on a dataset by training on the training set and evaluating their performance on the testing set. 
The final score of a model on a dataset is computed by averaging over the testing set scores for ten different stratified train-test splits of the dataset. 

\begin{figure*}
\begin{subfigure}{.5\textwidth}
    \centering
    \includegraphics[trim=300 165 300 160, clip, width=0.98\linewidth]{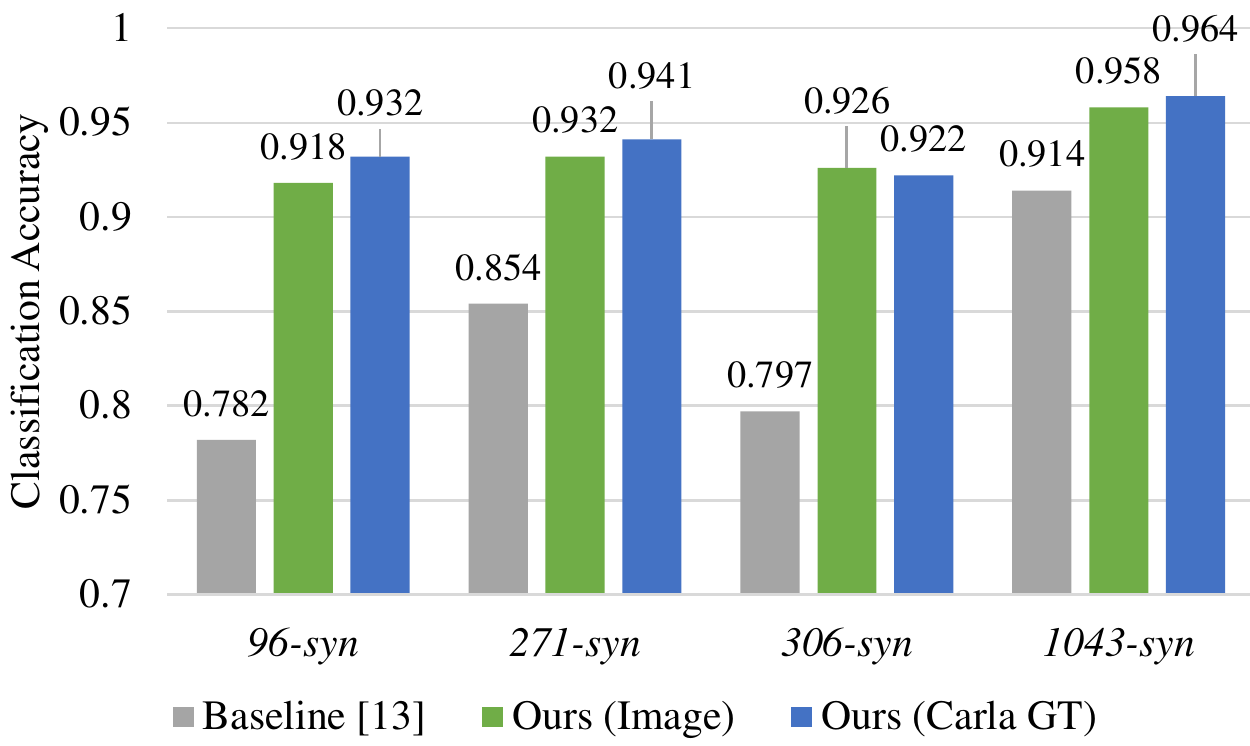}
    \label{fig:my_label1}
\end{subfigure}
\begin{subfigure}{.5\textwidth}
    \centering
    \includegraphics[trim=300 165 300 160, clip, width=0.98\linewidth]{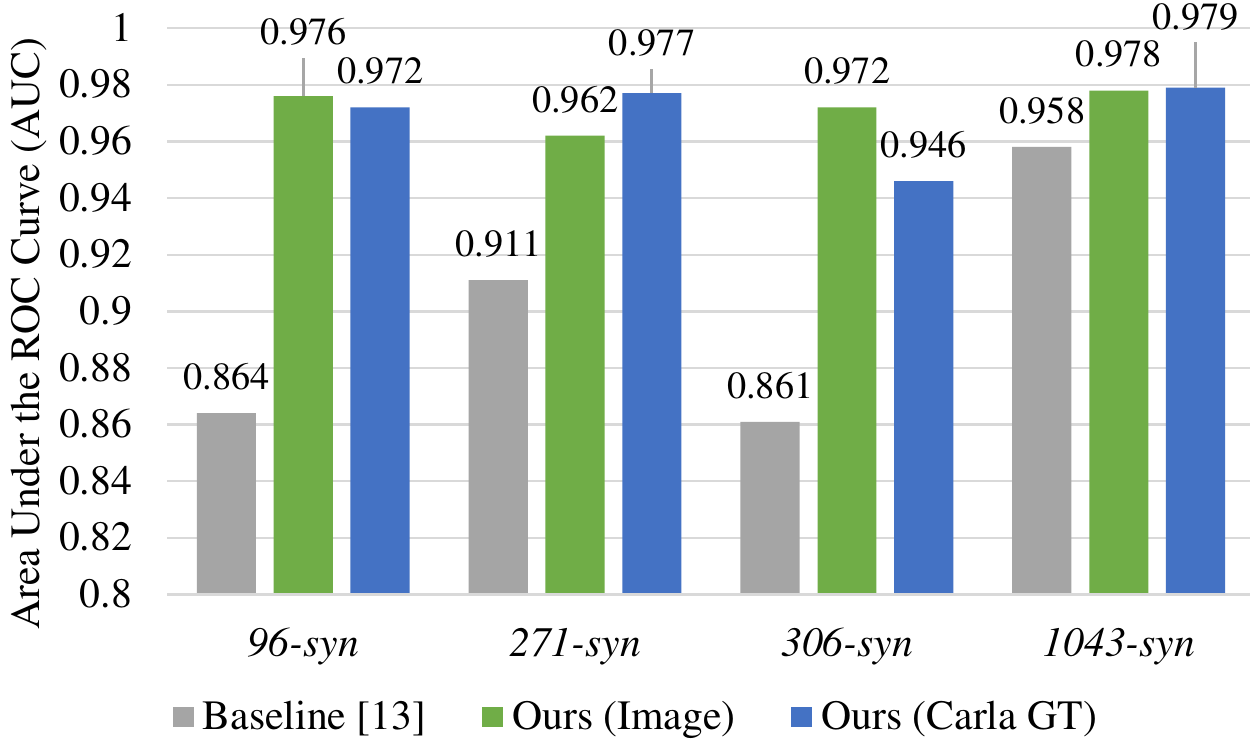}
    \label{fig:my_label2}
\end{subfigure}
    \caption{Accuracy and AUC comparison between our approaches (Real Image and Carla GT) and \cite{yurtsever2019risky} on different datasets. In these experiments, we trained the model using the hyper-parameter settings reported in Section~\ref{subsec:exp1}. }
    \label{fig:risk_result_1}
\end{figure*}

\subsection{Training and Model Specification}
\label{subsec:training}

Our models were implemented using \textit{PyTorch} and \textit{PyTorch-Geometric}~\cite{paszke2017automatic, Fey_Lenssen_2019}. 
We used the ADAM optimizer for the training algorithm.
We considered three learning rates: $\{0.0005, 0.0001, 0.00005\}$, and a weight decaying rate of $5\times 10^{-4}$ per epoch. 
We used a batch size of 16 sequences for each training epoch. 
In our experiments, we trained each model for 200 epochs.
Regarding the setting of hyper-parameters, we considered the options described in Section~\ref{subsec:model}.
To ensure a fair comparison between our model and the baseline model, we reported the model's performance with the lowest validation loss throughout the training for scoring.

All the experiments were conducted on a server with one NVIDIA TITAN-XP graphics card and one NVIDIA GeForce GTX 1080 graphics card.
For implementing the baseline model \cite{yurtsever2019risky}, we used the source code available at their open-source repository\footnote{\url{https://github.com/Ekim-Yurtsever/DeepTL-Lane-Change-Classification}}. 
Our model implementation is also available at \url{https://github.com/louisccc/av_av}.

\subsection{Experiments on Risk Assessment}
\label{subsec:exp1}
We evaluate each model's performance by measuring its classification accuracy and the Area Under the Curve (AUC) of the Receiver Operating Characteristic (ROC) for each dataset.
The classification accuracy is the ratio of the number of correct predictions on the test set of a dataset to the total number of samples in the testing set.
AUC, sometimes referred to as a balanced accuracy measure~\cite{sokolova2009systematic}, measures the probability that a binary classifier ranks a positive sample more highly than a random negative sample.
This is a more balanced measure for measuring accuracy, especially with imbalanced datasets (i.e. \textit{271-syn}, \textit{1043-syn}, \textit{571-honda}). 

From our experimentation, we found that the best option for the hyper-parameters of our model is a mini-batch size of 16 sequences, a learning rate of 0.00005, two MR-GCN layers with 100 hidden units, a SAGPool pooling layer with ratio 0.5, \textit{sum-pooling} for graph readout operation and \textit{LSTM-attn} for temporal modeling.

Figure~\ref{fig:risk_result_1} shows the comparison between our model's performance and the baseline model~\cite{yurtsever2019risky} for all the synthetic datasets. The results show that our approach consistently outperforms~\cite{yurtsever2019risky} across all the datasets in terms of both classification accuracy and AUC.
Particularly, on the \textit{1043-syn} dataset, our Image-based and GT pipelines outperform~\cite{yurtsever2019risky} in classification accuracy by 4.4\% and 5\% respectively (i.e., an accuracy of 95.8\% and 96.4\% compared to 91.4\% for the baseline).

We found that the performance difference between our approach and the baseline increased when the training datasets were smaller.
Figure~\ref{fig:risk_result_1} shows that the difference in the accuracy between our approach using the GT pipeline and the baseline~\cite{yurtsever2019risky} is 5\% for the \textit{1043-syn} dataset and 8.7\% for the  \textit{271-syn} dataset. 
This indicates that our approach can learn an accurate model even from a smaller dataset. 
We postulate this is a direct result of its use of a scene-graph based IR.

We also found that our approach performs better than the baseline on balanced datasets. 
Among the datasets used for evaluation of the models, the datasets \textit{271-syn} and \textit{306-syn} contain roughly the same number of clips but differ in the distribution of safe to risky lane changes (2.30:1 for \textit{271-syn} vs. 1:1 for \textit{306-syn}). 
We found that the performance difference between our image-based approach and the baseline on these datasets is 12.9\% on the \textit{306-syn} dataset compared to 7.8\% on the \textit{271-syn} dataset, indicating that our approach can discriminate between the two classes better than the baseline.

We also evaluated the contribution of each functional component in our proposed model by conducting an ablation study. 
The results of the study are shown in Table~\ref{tbl:ablation}.
From Table~\ref{tbl:ablation} we find that the simplest of the models, with no MR-GCN layer (replaced with an MLP layer) and a simple average of the embeddings in $\mathbf{h}_I$ for the temporal model (denoted as \textit{mean} in Table~\ref{tbl:ablation}), achieves a classification accuracy of 75\%.
We find that replacing \textit{mean} with an LSTM layer for temporal modeling yields a 10.5\% increase in performance.
We also find that including a single MR-GCN with 64 hidden units and \textit{sum-pooling} to the simplest model results in 14.8\% performance gain over the simplest model.
The performance gain achieved just by including the MR-GCN layer clearly suggests the effectiveness of modeling the relations between the objects.
Finally, we find that the model with one MR-GCN with 64 hidden units and \textit{sum-pooling} plus the LSTM layer for temporal modeling yields the maximum gain of 18.1\% over the simplest model.
These results clearly demonstrate the importance of each component in the model we propose.
\begin{table}
\begin{center}
\begin{tabular}{p{35pt} p{75pt} p{40pt} p{20pt} p{20pt}}
    \hline
     & Spatial Modeling & Temporal Modeling & Avr. Acc. & Avr. AUC\\
    \hline
    \hline
    \multirow{4}{8pt}{Ablation Study}&No MR-GCN & \textit{mean} & 0.762 & 0.823 \\\cline{2-5}
    &No MR-GCN & \textit{LSTM-last} & 0.867 & 0.929\\\cline{2-5}
    &1 MR-GCN & \textit{mean} & 0.910 & 0.960\\\cline{2-5}
    &1 MR-GCN & \textit{LSTM-last} & 0.943 & 0.977\\
    \hline
    \hline
    \multirow{4}{8pt}{Temporal Attention}&No MR-GCN & \textit{LSTM-last} & 0.867 & 0.929\\\cline{2-5}
    &No MR-GCN & \textit{LSTM-attn} & 0.868 & 0.928\\\cline{2-5}
    &1 MR-GCN & \textit{LSTM-last} & 0.943 & 0.977 \\\cline{2-5}
    &1 MR-GCN & \textit{LSTM-attn} & 0.950 & 0.977 \\
    \hline
    \hline
    \multirow{3}{8pt}{Spatial Attention}&1 MR-GCN & \textit{mean} & 0.910 & 0.960\\\cline{2-5}
    &1 MR-GCN, \textit{TopkPool} & \textit{mean} & 0.886 & 0.930 \\\cline{2-5}
    &1 MR-GCN, \textit{SAGPool} & \textit{mean} & 0.937 & 0.968 \\
    \hline
\end{tabular}
\end{center}
\caption{The results of the Carla GT approach on \textit{1043-syn} dataset with various spatial and temporal modeling settings. In these experiments, we used MR-GCN layers with 64 hidden units and \textit{sum-pooling} as the graph readout operation.}
\label{tbl:ablation}
\end{table}

\begin{figure}[!ht]
    \centering
    \includegraphics[trim=300 160 295 160, clip, width=\linewidth]{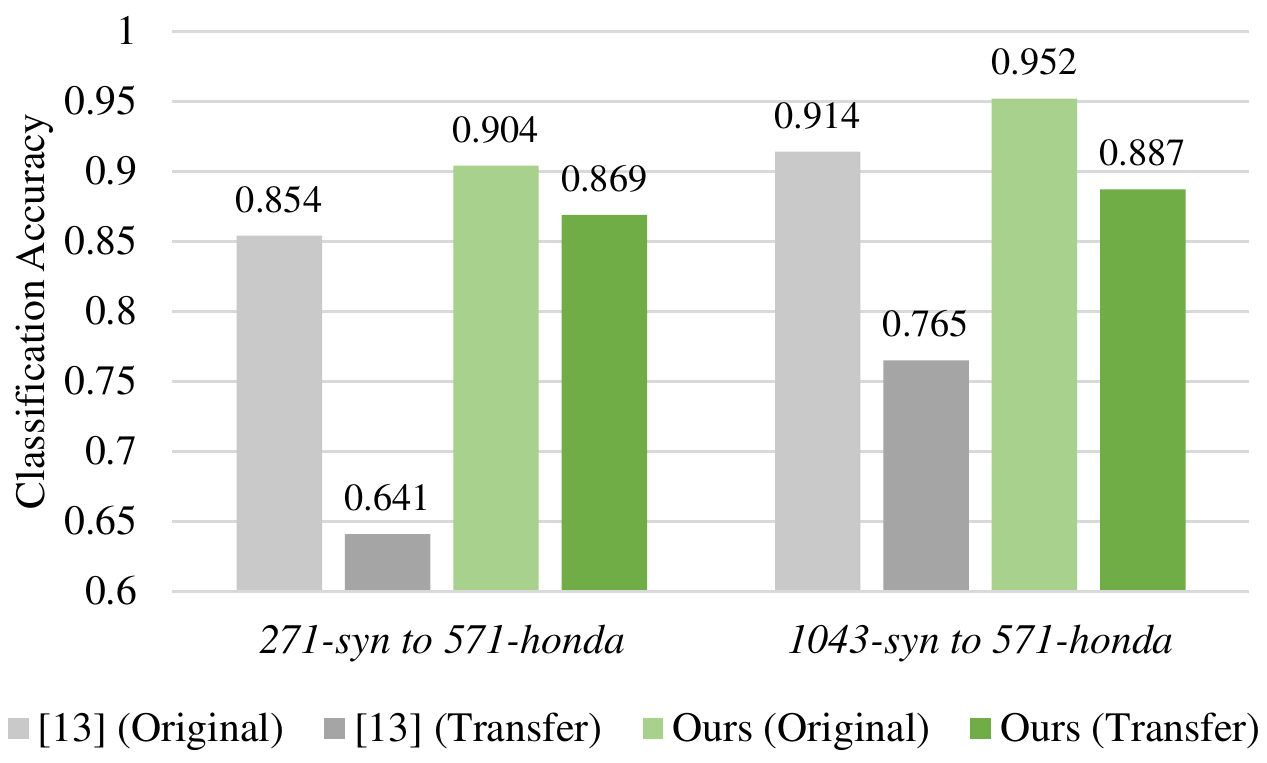}
    \caption{The results of comparing transferability between our Real Image model and \cite{yurtsever2019risky}. In this experiment, we trained our model using our best hyper-parameters on both \textit{271-syn} dataset and \textit{1043-syn} dataset.
    Then we tested the accuracy of our approach on both original dataset and \textit{571-honda} dataset. We followed the same procedure to train and test \cite{yurtsever2019risky}.}
    \label{fig:transfer}
\end{figure}

\subsection{Evaluation of Attention Mechanisms on Risk Assessment}
In this section, we evaluate the various attention components of our proposed model.
To evaluate the benefit of attention over the spatial domain, we tested our model with three different graph attention methods: no attention, \textit{SAGPool}, and \textit{TopkPool}.
To evaluate the impact of attention on the temporal domain, we tested our model with the following temporal models: \textit{mean}, \textit{LSTM-last}, and \textit{LSTM-attn}.
The detailed results that elucidate the effectiveness of these different attention mechanisms are presented in Table~\ref{tbl:ablation}.

For evaluating the benefits of graph attention, we start with an attention-free model: one MR-GCN layer with \textit{sum-pooling} + \textit{mean}.
In comparison, the model that uses \textit{SAGPool} for attention on the graph shows a 2.7\% performance gain over the attention-free model.
This indicates that the use of attention over both nodes and relations allows \textit{SAGPool} to better filter out irrelevant nodes from each \textit{scene-graph}. 
We found that the model using \textit{TopkPool} as the graph-attention layer became relatively unstable, resulting in a 2.4\% performance drop compared to the attention-free model.
This can be because \textit{TopkPool} ignores the relations of a node when calculating $\alpha$.
Another reason for this instability could be the random initialization of weights in \textit{TopkPool}, which can exponentially affect the overall performance as stated in~\cite{knyazev2019understanding}.

For evaluating the impact of attention on the temporal model, we evaluated the effects of adding a temporal attention layer to the following two models: (i) with no MR-GCN layers and no temporal attention and (ii) with one MR-GCN layer and no temporal attention.
Compared to the model with no MR-GCN layer and no temporal attention, the performance of the model with no MR-GCN and \textit{LSTM-attn} was found to be 0.1\% higher. 
We also found that adding \textit{LSTM-attn} to the model with one MR-GCN layer increases its performance by 0.7\% over the same model with no temporal attention.
These results demonstrate that the inclusion of temporal attention does improve the performance, though only marginally.
The reason why we only see a marginal improvement can be that the temporal attention layer is less relevant with the dataset that our model was trained on. 
When preparing these datasets, we manually removed the frames that are irrelevant to a lane change, exactly the set of frames that temporal attention would have given less attention to, thus minimizing its effect.

\begin{figure*}[!ht]
\centering
\includegraphics[width=1.0\linewidth]{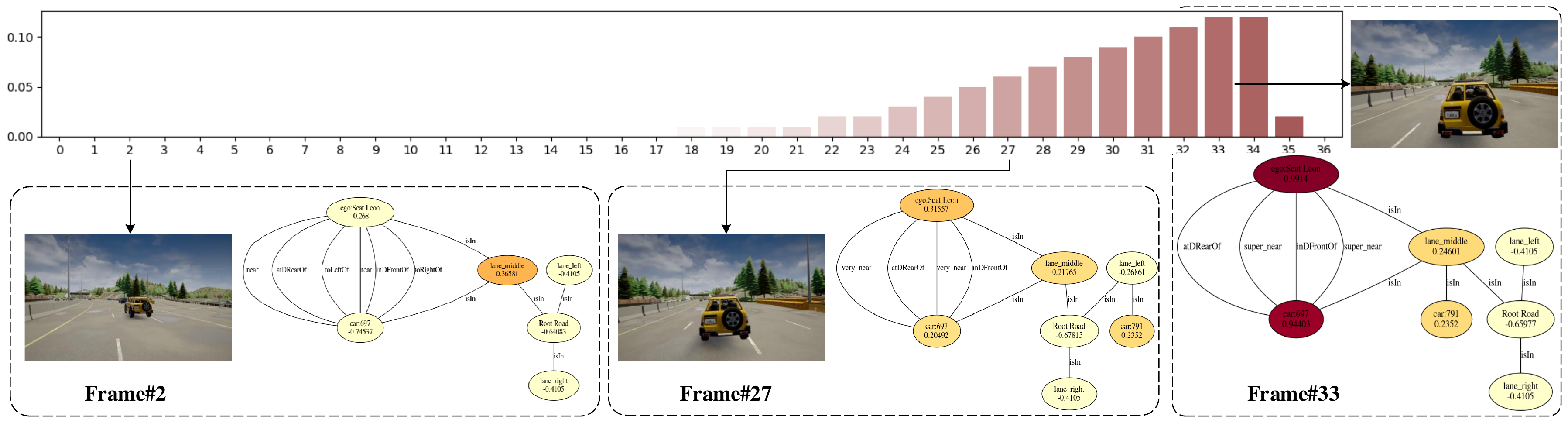}
\caption{The visualization of attention weights in both spatial ($\alpha$) and temporal ($\beta$) domains using a risky lane changing clip as an example. We used a gradient color from light yellow to red for visualizing each node's projection score that indicates its importance in calculating a \textit{scene-graph} embedding. We also used a gradient colored (white to red) bar chart to visualize the temporal attention coefficients $\beta_1, \beta_2, ... \beta_{36}$ used for calculating the context vector $c$.}
\label{fig:attention}
\end{figure*}

The primary benefit of using the graph-attention and the temporal attention is that it improves the explainability of the model's risk assessment.
We demonstrate this capability using the visualization of both graph attention and temporal attention provided in Figure~\ref{fig:attention}.
Figure~\ref{fig:attention} shows the trend of the attention scores $\beta_1, \beta_2, ... \beta_T$ for a risky lane changing clip. 
Intuitively, the frame with a higher attention score $\alpha_t$ contributes more to the context vector $c$ (shown in Equation~\ref{eq:temporalattn}), thus playing a more critical role in calculating $h_{G_t}$ and contributing to the final risk assessment decision. 
In this risky lane changing example, the temporal attention scores progressively increase between frames 19 and 32 during the lane change; and the highest frame attention weights appear in frames 33 and 34, which are the frames immediately before the collision occurs. 
Figure~\ref{fig:attention} also shows the projection scores for the node attention layer, where a higher score for a node indicates that it contributes more to the final decision of risk assessment.
As shown in this example, as the ego car approaches the yellow vehicle, the node attention weights for the ego car and the yellow vehicle are increased proportionally to the scene's overall risk. 
In the first few frames, the risk of collision is low; thus the node attention weights are low; however, in the last few frames, a collision between these two vehicles is imminent; thus the attention weights for the two cars are much higher than for any other nodes in the graph.
This example clearly demonstrates our model's capability to pinpoint the critical factors in a \textit{scene-graph} that contributed to its risk assessment decision. 
This capability can be valuable for debugging edge cases at design time, thus reducing the chances of ADS making unexpected, erroneous decisions in real-world scenarios and improving human trust in the system.

\subsection{Transferability from Virtual To Real Driving}
In this section, we demonstrate our approach's capability to effectively transfer the knowledge learned from a simulated dataset to a real-world dataset.
To demonstrate this capability, we use the model weights, and parameters learned from training on the \textit{271-syn} dataset or the \textit{1043-syn} dataset directly for testing on the real-world driving dataset: \textit{571-honda}. 
We also compare the transferability of our model with that of the baseline method~\cite{yurtsever2019risky}.
The results are shown in Figure~\ref{fig:transfer}.

As expected, the performance of both our approach and the baseline degrades when tested on \textit{571-honda} dataset. 
However, as Figure~\ref{fig:transfer} shows, the accuracy of our approach only drops by 6.7\% and 3.5\% when the model is trained on \textit{271-syn} and \textit{1043-syn}, respectively, while the baseline's performance drops drastically by a much higher 21.3\% and 14.9\%, respectively. 
The results categorically show that our proposed model can transfer knowledge more effectively than the baseline.

%% file: sections/5_conclusion.tex
\section{Conclusion}
\label{sec:conclusion}
Subjective risk assessment is a challenging, safety-critical problem that requires a good semantic understanding of many possible road scenarios. 
Our results show that our scene-graph augmented approach outperforms state-of-the-art techniques at risk assessment tasks in terms of accuracy (95.8\% vs. 91.4\%) and AUC (0.978 vs. 0.958).
We also show that our approach can learn with much less training data than these techniques, as our approach achieves 91.8\% accuracy on the \textit{96-syn} dataset compared to 78.2\% accuracy achieved by \cite{yurtsever2019risky}.
Additionally, our results show that our approach can better transfer knowledge gained from simulated datasets to real-world datasets (5.0\% avg. acc. drop for our approach vs. 18.1\% avg. acc. drop for \cite{yurtsever2019risky}).
We also show that the spatial and temporal attention components used in our approach improve both its performance and its explainability.

%% file: sections/6_acknowledgement.tex
\section*{Acknowledgment}
The authors would like to show gratitude to the colleagues of the Embedded \& Cyber-Physical Systems (AICPS) Lab, from University of California, Irvine, particularly to Emily Sing Yen Lam and Aung Myat Thu for the contribution during the course of this research. 
This work was partially supported by NSF under award CMMI-1739503 and the award ECCS-1839429. 
Any opinions, findings, conclusions, or recommendations expressed in this paper are those of the authors and do not necessarily reflect the views of the funding agency.

%% file: sections/7_IEEE_author_bios.tex
\begin{IEEEbiography}[
    {\includegraphics[width=1in,height=1.25in,clip,keepaspectratio]
{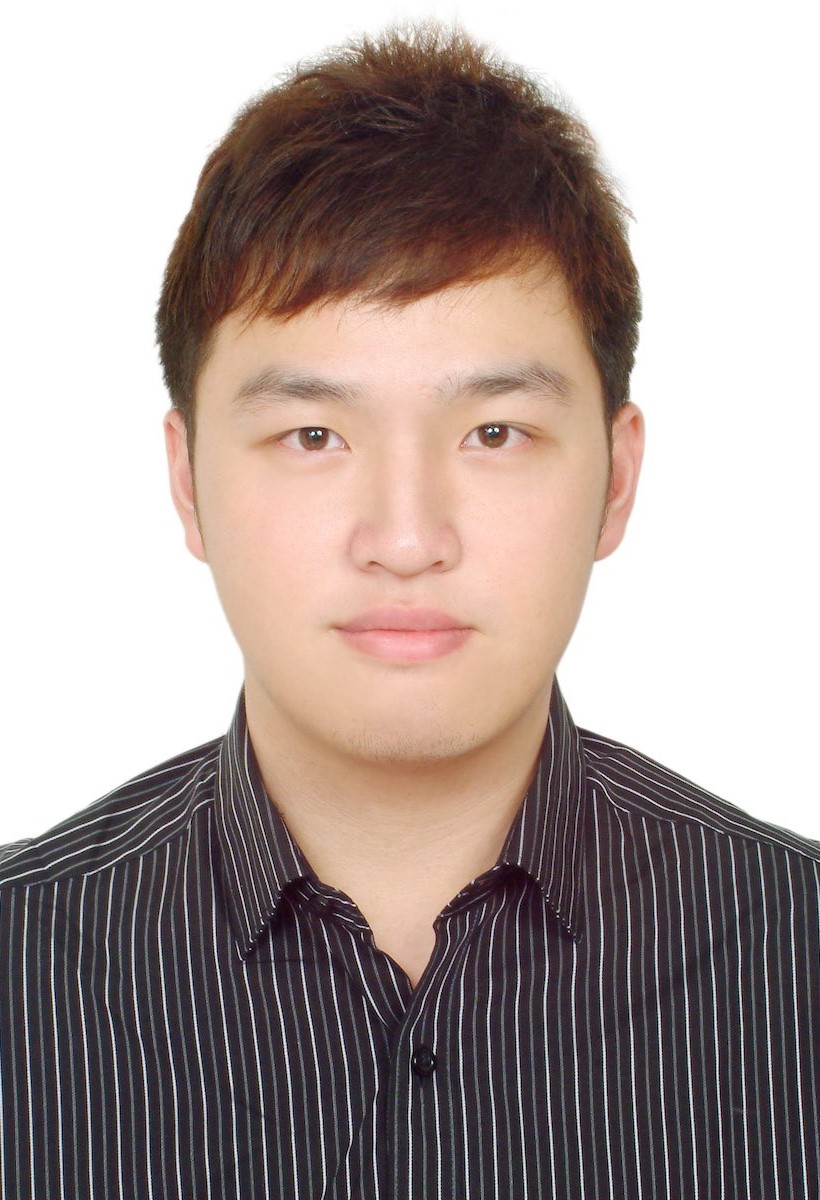}}]
{Shih-Yuan Yu} received the B.S. and M.S. degrees in Computer Science and Information Engineering from the National Taiwan University (NTU) in 2014. He worked at MediaTek for 4 years. Currently he is a Ph.D. student in the University of California, Irvine.  
Now his research interests are about design automation of embedded systems using data-driven system modeling approaches.
It covers incorporating machine learning methods to identify potential security issues in systems.
\end{IEEEbiography}

\begin{IEEEbiography}[
    {\includegraphics[width=1in,height=1.25in,clip,keepaspectratio]
{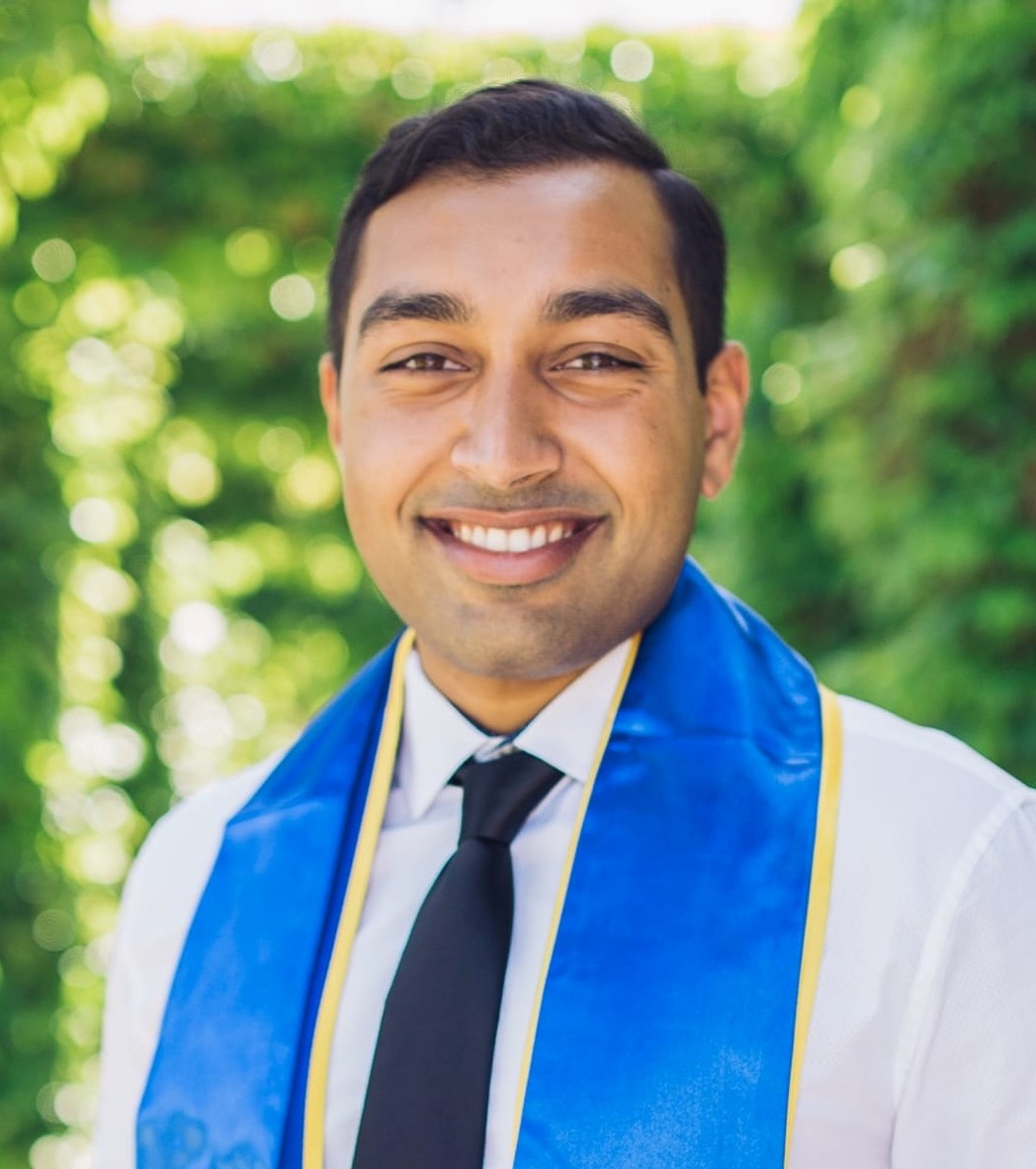}}]
{Arnav Vaibhav Malawade} received a B.S. in Computer Science and Engineering from the University of California Irvine (UCI) in 2018. He is currently an M.S./Ph.D. Student studying  Computer Engineering at UCI under the supervision of Professor Mohammad Al Faruque. His research interests include the design and security of cyber-physical systems in connected/autonomous vehicles, manufacturing, IoT, and healthcare.
\end{IEEEbiography}

\begin{IEEEbiography}[{\includegraphics[width=1in,height=1.25in,clip,keepaspectratio]{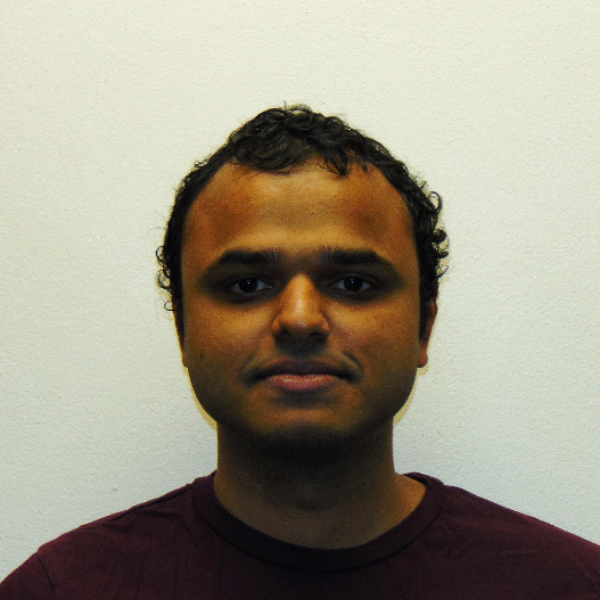}}]{Deepan Muthirayan}
is currently a Post-doctoral Researcher in the department of Electrical Engineering and Computer Science at University of California at Irvine. He obtained his Phd from the University of California at Berkeley (2016) and B.Tech/M.tech degree from the Indian Institute of Technology Madras (2010). His doctoral thesis work focussed on market mechanisms for integrating demand flexibility in energy systems. Before his term at UC Irvine he was a post-doctoral associate at Cornell University where his work focussed on online scheduling algorithms for managing demand flexibility. His current research interests include control theory, machine learning, topics at the intersection of learning and control, online learning, online algorithms, game theory, and their application to smart systems.
\end{IEEEbiography}

\begin{IEEEbiography}
[{\includegraphics[width=1in,height=1.25in,clip,keepaspectratio]{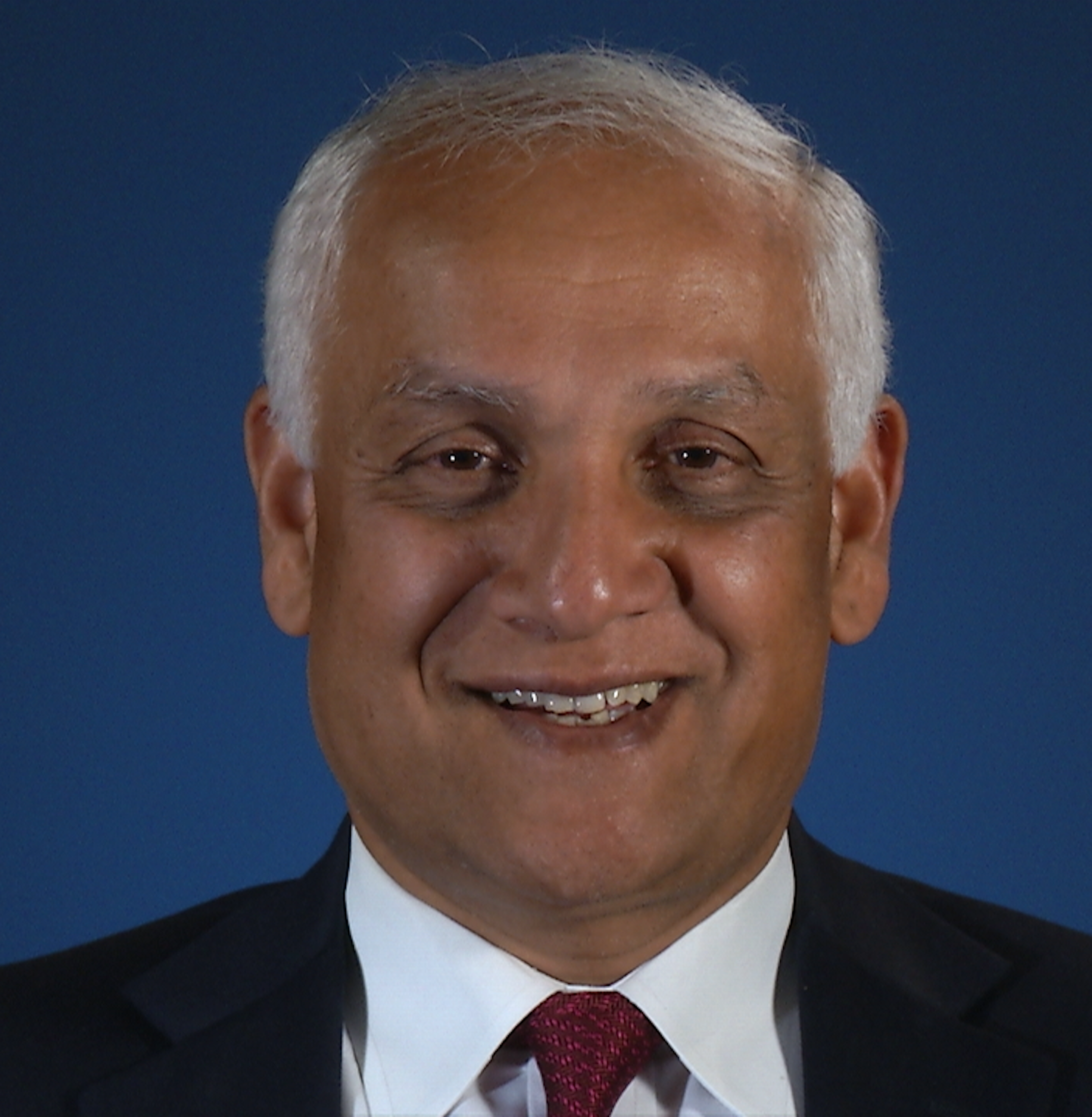}}]
{Pramod Khargonekar} received B. Tech. Degree in electrical engineering in 1977 from the Indian Institute of Technology, Bombay, India, and M.S. degree in mathematics in 1980 and Ph.D. degree in electrical engineering in 1981 from the University of Florida, respectively. He was Chairman of the Department of Electrical Engineering and Computer Science from 1997 to 2001 and also held the position of Claude E. Shannon Professor of Engineering Science at The University of Michigan.  From 2001 to 2009, he was Dean of the College of Engineering and Eckis Professor of Electrical and Computer Engineering at the University of Florida till 2016. After serving briefly as Deputy Director of Technology at ARPA-E in 2012-13, he was appointed by the National Science Foundation (NSF) to serve as Assistant Director for the Directorate of Engineering (ENG) in March 2013, a position he held till June 2016. Currently, he is Vice Chancellor for Research and Distinguished Professor of Electrical Engineering and Computer Science at the University of California, Irvine. His research and teaching interests are centered on theory and applications of systems and control. He has received numerous honors and awards including IEEE Control Systems Award, IEEE Baker Prize, IEEE CSS Axelby Award, NSF Presidential Young Investigator Award, AACC Eckman Award, and is a Fellow of IEEE, IFAC, and AAAS.
\end{IEEEbiography}

\begin{IEEEbiography}
[{\includegraphics[width=1in,height=1.25in,clip,keepaspectratio]
{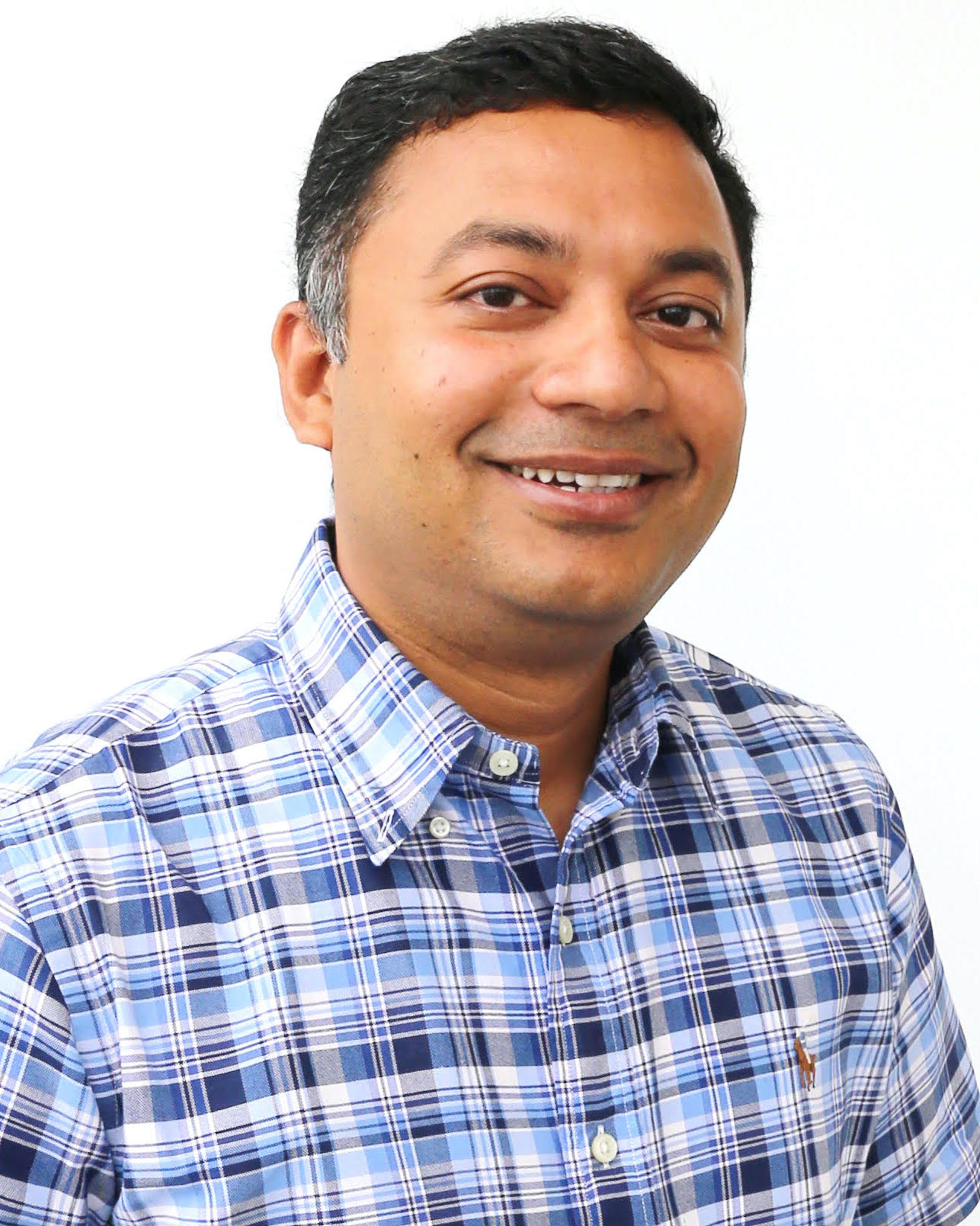}}]
{Mohammad Abdullah Al Faruque} (M’06, SM’15) received his B.Sc. degree in Computer Science and Engineering (CSE) from Bangladesh University of Engineering and Technology (BUET) in 2002, and M.Sc. and Ph.D. degrees in Computer Science from Aachen Technical University and Karlsruhe Institute of Technology, Germany in 2004 and 2009, respectively.
He is currently with the University of California Irvine (UCI) as an Associate Professor and Directing the Embedded and Cyber-Physical Systems Lab. He served as an Emulex Career Development Chair from October 2012 till July 2015. Before, he was with Siemens Corporate Research and Technology in Princeton, NJ as a Research Scientist. His current research is focused on the system-level design of embedded and Cyber-Physical-Systems (CPS) with special interest in low-power design, CPS security, data-driven CPS design, etc.
He is an ACM senior member. He is the author of 2 published books. Besides many other awards, he is the recipient of the School of Engineering Mid-Career Faculty Award for Research 2019, the IEEE Technical Committee on Cyber-Physical Systems Early-Career Award 2018, and the IEEE CEDA Ernest S. Kuh Early Career Award 2016. He is also the recipient of the UCI Academic Senate Distinguished Early-Career Faculty Award for Research 2017 and the School of Engineering Early-Career Faculty Award for Research 2017. Besides 120+ IEEE/ACM publications in the premier journals and conferences, he holds 8 US patents.
\end{IEEEbiography}